\documentclass[10pt]{article}
\usepackage[utf8]{inputenc}
\usepackage[T1]{fontenc}
\usepackage{mathptmx}
\usepackage[letterpaper, margin=1in]{geometry}
\usepackage{amsmath, amssymb}
\usepackage{graphicx}
\DeclareGraphicsExtensions{.pdf,.png,.jpg}
\usepackage{booktabs}
\usepackage{xcolor}
\usepackage{caption}
\usepackage{subcaption}
\usepackage{tabularx}
\usepackage{setspace}
\usepackage[super,sort&compress]{natbib}
\usepackage{hyperref}
\newcommand{\expertllm}{\texttt{deepseek.v3.2}}
\begin{document}

\begin{center}
{\fontsize{14}{16.8}\selectfont \textbf{Enhancing In-Hospital Mortality Prediction Using\\ Multi-Representational Learning with LLM-Generated Expert Summaries}}
\end{center}

\vspace{6pt}

\begin{center}
{\fontsize{12}{14.4}\selectfont \textbf{%
{Harshavardhan Battula, MS}$^{1}$,
{Jiacheng Liu, MS}$^{1}$,
{Jaideep Srivastava, PhD}$^{1}$
}}\\[4pt]
{\fontsize{10}{12}\selectfont
$^{1}${Department of Computer Science and Engineering, University of Minnesota Twin Cities, Minneapolis, MN, USA}\\
{battu018@umn.edu}%
}
\end{center}

\vspace{6pt}
\begin{center}
{\small
\textbf{Corresponding author:} Harshavardhan Battula, Department of Computer Science and Engineering, University of Minnesota Twin Cities, 200 Union Street SE, Minneapolis, MN 55455, USA. Email: battu018@umn.edu\\[3pt]
\textbf{Article type:} Brief Communication \quad
\textbf{Word count (body):} 2{,}000 \quad
\textbf{Abstract word count:} 128 \quad
\textbf{Tables:} 2 \quad \textbf{Figures:} 2 \quad \textbf{Supplementary tables:} 5}
\end{center}

\vspace{12pt}

\noindent\textbf{Abstract}

\vspace{4pt}

\noindent\textit{%
To evaluate a multi-representational framework in which large language model (LLM)-generated expert summaries of intensive care unit (ICU) notes are fused with physiology for in-hospital mortality (IHM) prediction, and to determine how much of the resulting gain is non-redundant with the notes themselves. Using MIMIC-III (19,211 first ICU stays, 12.83\% mortality), we encoded 48-hour physiology, clinical notes, and LLM summaries generated under a prompt forbidding prognostication, then fused them. Redundancy was assessed by ridge recoverability, linear probes, and retrained ablations substituting a note-orthogonal residual or a patient-shuffled summary embedding. On 3,843 held-out stays, fusion reached AUPRC 0.4977/AUROC 0.8429 versus 0.3625/0.7770 for physiology alone. Ridge regression from note embeddings explained 40.8\% of summary-embedding variance. The note-orthogonal residual retained a minority of the gain (+0.0258 AUPRC, 95\% CI 0.005--0.047; 28\%), while patient-shuffled summaries fell below the reference. Summaries improve prediction patient-specifically, but predominantly by reorganizing information the notes already contain.%
}

\vspace{6pt}


\vspace{8pt}

\noindent\textbf{Background and Significance}

\vspace{4pt}

Predicting in-hospital mortality (IHM) for intensive care unit (ICU) patients supports early intervention and resource planning. Most models use structured physiology alone,\textsuperscript{\ref{harutyunyan2019multitask},\ref{ghassemi2015multivariate},\ref{suresh2018learning},\ref{song2018attend}} discarding the clinical narrative that transformer encoders pre-trained on clinical text\textsuperscript{\ref{alsentzer2019publicly},\ref{lee2020biobert}} have since made tractable as model input.

A newer line of work interposes a large language model (LLM) between the notes and the predictor, using it to generate a structured summary that is then encoded and fused with physiology. Because using LLMs as predictors directly raises unresolved calibration problems,\textsuperscript{\ref{bommasani2022foundation},\ref{tian2023justask}} restricting the LLM to \emph{representation} is attractive: the downstream network retains ownership of the probability and can be calibrated by standard means. Reported gains from this design are consistent, but its mechanism is untested.

The summary is generated deterministically from the notes, yet the LLM's parameters also supply external medical knowledge, so a gain could arise from three sources: the LLM surfacing content the note encoder fails to reach (\emph{reorganization}); the summary embedding reaching a region of representation space the note embedding does not (\emph{complementarity}); or the LLM applying knowledge absent from the record. Our design separates the note-recoverable component from the remainder, not the latter two from each other. That first split alone, however, already determines where effort belongs: a better text encoder, or a better prompt and LLM. To our knowledge no prior evaluation of LLM-derived clinical representations has attempted it.

\vspace{8pt}

\noindent\textbf{Objective}

\vspace{4pt}

We quantify how much of an expert summary's contribution to IHM prediction is non-redundant with the notes it was written from, using representational geometry, linear recoverability with label probes, and retrained ablations.

\vspace{8pt}

\noindent\textbf{Materials and Methods}

\vspace{4pt}

\noindent\textit{Cohort}

\vspace{4pt}

We used MIMIC-III.\textsuperscript{\ref{johnson2016mimic}} Inclusion criteria, in order: first ICU stay per patient; age $\geq$18 at admission (ages over 89 clipped to 90); ICU length of stay exceeding the 48-hour observation window; survival beyond the prediction point; at least one usable in-window note; and at least one in-window physiological measurement. The label is \texttt{HOSPITAL\_EXPIRE\_FLAG}.

Three leakage controls warrant emphasis. Stays in which death occurred inside the 48-hour window were excluded, since for these patients the outcome is already determined at the prediction point and the notes describe it. Positive labels with a null \texttt{DEATHTIME} were excluded, because the death cannot be shown to postdate that point. Notes whose \texttt{STORETIME} fell after hour 48 were excluded: radiology reports and addenda are routinely charted inside the window but signed days later, and can describe the remainder of the admission.

The first four criteria left 20,226 stays (13.07\% mortality). Requiring a usable note removed 934 stays, 178 of them deaths; requiring a physiological measurement removed a further 81, all survivors. The analytic cohort is \textbf{19,211 stays with 2,465 deaths (12.83\%)}: median age 66.9 years (18--90), 56.2\% male, 70.9\% White, median ICU length of stay 3.98 days (Supplementary Table S1). Splits were stratified 60/20/20 (11,526/3,842/3,843), each at 12.83\% mortality, and are patient-disjoint by construction because one stay is retained per patient.

\vspace{4pt}

\noindent\textit{Representations}

\vspace{4pt}

Ten physiological variables (blood pressures, heart rate, temperature, respiratory rate, SpO$_2$, FiO$_2$, pH, glucose) were extracted from 61 item IDs following the Harutyunyan et al.\ benchmark,\textsuperscript{\ref{harutyunyan2019multitask}} unit-harmonized, range-checked, and averaged within each of 48 hours. Missing hours were forward-filled then imputed with benchmark normal values, with a parallel binary mask so the network can distinguish ``normal'' from ``never measured'' (overall mask density 0.635, but 0.14 for pH and 0.16 for FiO$_2$; Supplementary Table S2). A single-layer LSTM\textsuperscript{\ref{hochreiter1997long}} yields the 256-dimensional $H_T$.

Discharge summaries, error-flagged notes, and notes lacking \texttt{CHARTTIME} were excluded, leaving 200,145 in-window notes across the analytic cohort (10.4 per stay; Supplementary Table S3). Each note was encoded independently with Bio\_ClinicalBERT\textsuperscript{\ref{alsentzer2019publicly}} and mean-pooled over the attention mask (768-d); the 32 most recent notes per stay were encoded (195,699 notes, 97.8\%; mean 10.2 per stay). The note representation applies exponential temporal decay,
\begin{equation}
U_T = \frac{1}{M}\sum_{i:\,\mathrm{CT}(i)\leq T} U_i\,\exp\!\big(-\lambda\,(T-\mathrm{CT}(i))\big),
\end{equation}

with $M$ the number of in-window notes and $\lambda$ tuned on validation AUPRC using the notes-only configuration.

For the expert branch, \emph{all} in-window notes were concatenated chronologically, each prefixed with category and hour offset, and truncated to 4,096 ClinicalBERT tokens by discarding the \emph{oldest} tokens, retaining the hours nearest the prediction point. \expertllm{}, self-hosted on a single node of eight NVIDIA H100 GPUs and served through a locally deployed OpenAI-compatible endpoint, produced a summary under six fixed headings (presentation, active problems, trajectory, organ support, concerning findings, clinical gestalt), constrained to use only information present in the notes, to leave de-identification placeholders untouched, and not to state, predict, or imply survival or death, nor give any numeric risk estimate. Generation used temperature 0 and a 700-token cap; median summary length was 264 words (p05 141, p95 344), encoded by the same encoder to give $V$.

\begin{figure}[!t]
\centering
\includegraphics[width=0.85\textwidth]{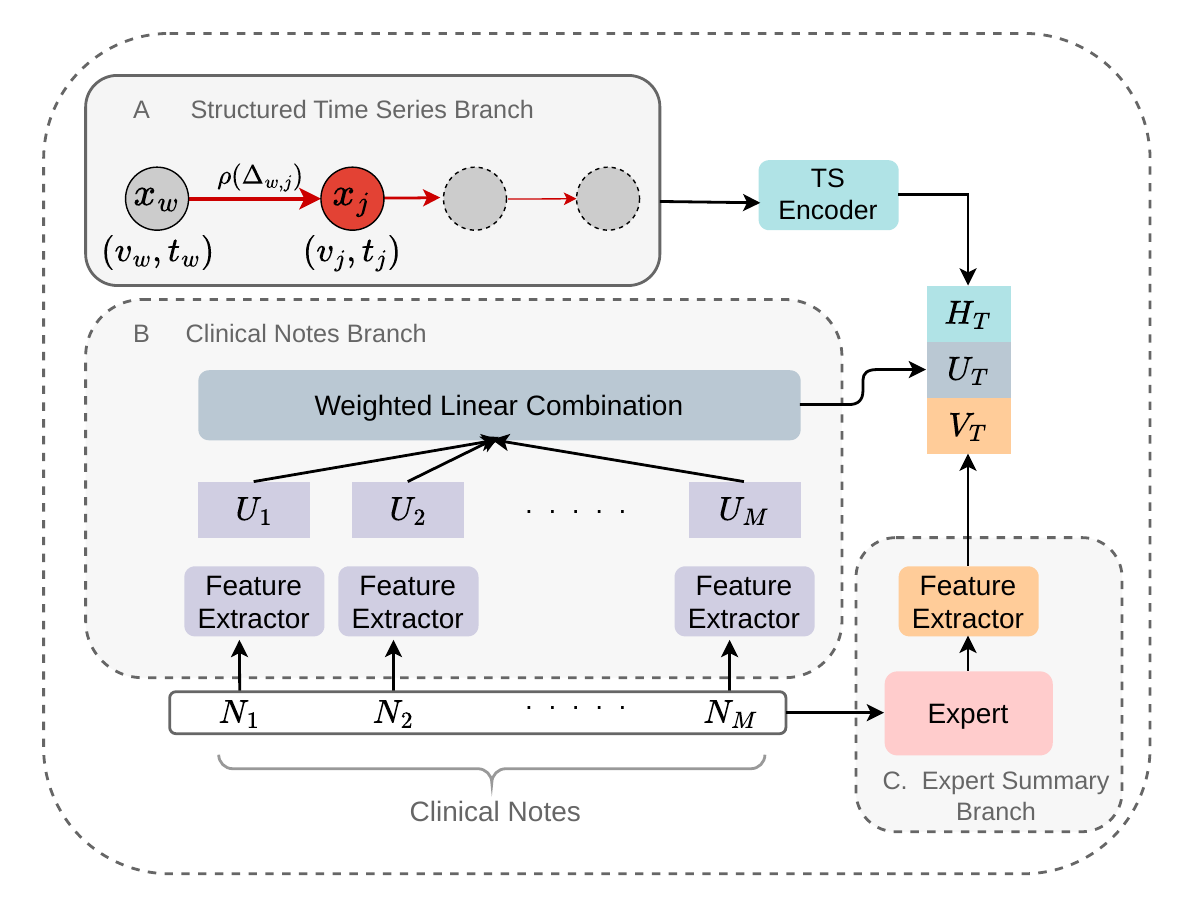}
\caption{Multi-representational learning framework. Hourly physiology is encoded by an LSTM to $H_T$ (256-d); clinical notes are encoded per note by Bio\_ClinicalBERT and pooled to $U_T$ (768-d); all in-window notes are summarized by an LLM under a prompt forbidding prognostication and the summary is encoded to $V$ (768-d). Enabled branches are concatenated and mapped to a mortality probability by a single linear layer.}
\label{fig:framework}
\end{figure}

Fusion concatenates the enabled branches, applies dropout, and maps to a logit through a single linear layer, $1{,}792\rightarrow1$ for the full model (Figure~\ref{fig:framework}). Training used unweighted binary cross-entropy, dropout 0.3, Adam (lr $1\times10^{-4}$, weight decay $1\times10^{-5}$), batch 32, gradient clipping 5.0, and up to 50 epochs with early stopping on validation loss (patience 5), seed 42; physiological standardization statistics were fit on training rows only.

\vspace{4pt}

\noindent\textit{Redundancy analysis}

\vspace{4pt}

All quantities were computed on the held-out test split. \textbf{Geometry:} cosine similarity of matched versus mismatched $(V_i,U_j)$ pairs and linear CKA, both after subtracting the training-split mean; and canonical correlations on the top 50 principal components of each representation, with PCA bases fit on training rows but canonical directions solved on the test projections, so these are in-sample and upward-biased. \textbf{Recoverability:} ridge regression from standardized $U_T$ to $V$ ($\alpha$ selected on validation $R^2$), yielding $V_\perp = V - \mathbb{E}[V\mid U_T]$, with training-row predictions cross-fitted over 5 folds so the residual is out-of-sample everywhere; label signal was read out with $L_2$-regularized logistic probes ($C$ selected on validation AUROC). \textbf{Ablation:} the fusion architecture was retrained unchanged (same splits, hyperparameters, seed), substituting for the expert branch (i) train-standardized $V$, (ii) $V_\perp$, and (iii) $V$ permuted across patients \emph{within each split}, preserving each split's marginal distribution and class balance while destroying the patient-to-summary pairing.

Confidence intervals are 1,000-sample percentile bootstraps; comparisons use a paired bootstrap on identical resampled rows, unadjusted for multiplicity. Structural invariants (Supplementary Table S5) are re-checked before any metric is reported.

\vspace{8pt}

\noindent\textbf{Results}

\vspace{4pt}

Validation AUPRC for the notes-only model fell monotonically across the decay grid (0.3133, 0.3056, 0.2326, 0.1733 at $\lambda=0$, 0.01, 0.05, 0.10), so $\lambda=0$ was selected. We do not read this as evidence against temporal decay: the normalizer divides by note count rather than by the sum of weights, so larger $\lambda$ shrinks $\|U_T\|$ rather than reweighting it. A weight-normalized variant is untested.

In Table~\ref{tab:results}, the full multi-representational model reached AUPRC 0.4977 and AUROC 0.8429, 37.3\% and 8.5\% above the physiology-only baseline. Clinical notes alone were the weakest configuration on both metrics; expert summaries alone matched physiology on AUPRC and were numerically higher on AUROC, though the intervals overlap and no paired test was performed.

\begin{table}[ht]
\centering
\caption{In-hospital mortality prediction on the held-out test set ($n$=3,843; 493 deaths; 12.83\% prevalence). Percentage changes are relative to the physiology-only baseline. All configurations share architecture, splits, hyperparameters, and seed; $\lambda=0$. The final row standardizes the expert embedding on training statistics and is the recommended configuration.}
\label{tab:results}
\small
\begin{tabular}{lcccc}
\toprule
\textbf{Model} & \textbf{AUROC (95\% CI)} & \textbf{$\Delta$\%} & \textbf{AUPRC (95\% CI)} & \textbf{$\Delta$\%} \\
\midrule
Physiology only & 0.7770 [0.756, 0.800] & -- & 0.3625 [0.324, 0.408] & -- \\
Clinical notes only & 0.7473 [0.726, 0.768] & $-$3.83 & 0.3010 [0.262, 0.342] & $-$16.95 \\
Expert summary only & 0.7969 [0.776, 0.816] & $+$2.57 & 0.3626 [0.320, 0.409] & $+$0.04 \\
Physiology $+$ notes & 0.8000 [0.780, 0.821] & $+$2.96 & 0.4054 [0.362, 0.450] & $+$11.85 \\
Physiology $+$ notes $+$ summary & 0.8231 [0.805, 0.843] & $+$5.94 & 0.4550 [0.409, 0.501] & $+$25.52 \\
\midrule
\textbf{$+$ summary, standardized}$^{\dagger}$ & \textbf{0.8429} & \textbf{$+$8.48} & \textbf{0.4977} & \textbf{$+$37.30} \\
\bottomrule
\multicolumn{5}{p{0.93\textwidth}}{\footnotesize Intervals are 1,000-sample percentile bootstraps. $^{\dagger}$Absolute percentile intervals were computed for the five pre-specified configurations; the standardized variant is characterized instead by its paired-bootstrap gain over physiology $+$ notes: $+$0.0429 AUROC (95\% CI 0.029--0.058) and $+$0.0928 AUPRC (0.065--0.121). No paired test against the physiology-only baseline was performed, so the $\Delta$\% column for this row is descriptive.}
\end{tabular}
\end{table}

\begin{figure}[!t]
\centering
\includegraphics[width=\textwidth]{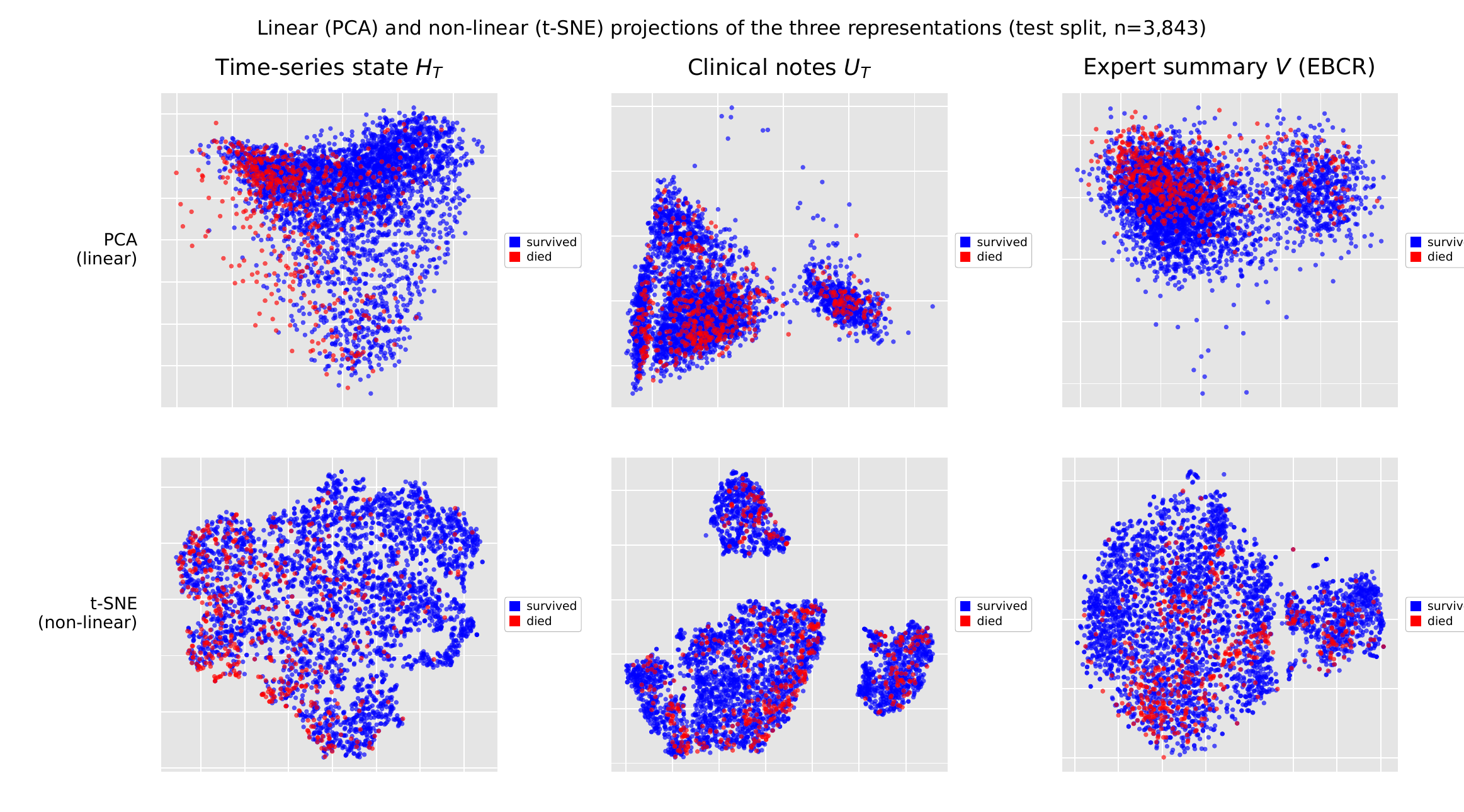}
\caption{Linear (PCA, top) and non-linear (t-SNE, bottom) projections of the three representations taken from the trained joint model on the test split ($n$=3,843; t-SNE with perplexity 30, PCA initialization, seed 42). Left, LSTM state $H_T$; center, pooled clinical-note embedding $U_T$; right, expert summary embedding $V$. Red denotes in-hospital death. Shown for qualitative illustration only: because these representations are optimized for the outcome, apparent class separation is not evidence about intrinsic information content, for which we rely on the probes in Table~\ref{tab:redundancy}.}
\label{fig:tsne}
\end{figure}

Standardizing $V$ was worth a further 0.0198 AUROC and 0.0427 AUPRC over the unstandardized fusion and was also superior on validation (AUPRC 0.4997 vs 0.4640), so this preference is not a test-set selection; a single linear head evidently does not absorb Bio\_ClinicalBERT's large common mean vector and heterogeneous per-dimension scales. Figure~\ref{fig:tsne} shows the three representations qualitatively.

\vspace{4pt}

\noindent\textit{Is the summary new information?}

\vspace{4pt}

\textbf{Geometry.} Matched pairs $(V_i,U_i)$ had mean cosine 0.2734 (SD 0.1322) against 0.0025 for mismatched pairs; without train-centering the matched cosine is 0.9371, an artifact of the shared mean vector. Retrieval was far above chance but not deterministic: querying each $V_i$ against all 3,843 note vectors recovered the correct stay at rank 1 in 7.1\% of cases (chance 0.03\%) and within the top 5 in 16.2\% (chance 0.13\%), with median rank 82; the true match falls in the top 2\% of candidates. Linear CKA was 0.5621 and the leading canonical correlation 0.927, with only 2 of 50 directions exceeding 0.9 (Supplementary Table S4). The pairing is strongly patient-identifying, yet the two vectors are not near-duplicates.

\textbf{Recoverability and probes.} Ridge regression from $U_T$ to $V$ achieved test $R^2=0.4078$; adding physiological summary statistics raised this only to 0.4120. Roughly 41\% of the variance of the expert embedding is a linear function of the pooled note embedding, and the cross-fitted residual retained 77.0\% of the norm of centered $V$. Table~\ref{tab:redundancy} shows most of the \emph{label} signal in $V$ nonetheless lives in its note-explainable part: $\hat{V}$ alone probes at AUROC 0.8157, close to $V$'s 0.8293, while $V_\perp$ reaches 0.6906. Concatenating both representations improved on $V$ alone by only 0.005 AUROC and 0.009 AUPRC, so at the level of linear read-out the summary embedding nearly subsumes the notes.

\begin{table}[ht]
\centering
\caption{Redundancy of the expert summary embedding $V$ with respect to the pooled note embedding $U_T$, test split. \textbf{Panel A}: $L_2$-regularized logistic probes on train-standardized inputs, regularization selected on validation AUROC. \textbf{Panel B}: the fusion architecture retrained with three substitutions for the expert branch; deltas are paired bootstraps (1,000 resamples) against the physiology-plus-notes reference, unadjusted for multiplicity. $\hat{V}=\mathbb{E}[V\mid U_T]$ from cross-fitted ridge regression ($\alpha$=1,000; test $R^2$=0.4078); $V_\perp = V-\hat{V}$.}
\label{tab:redundancy}
\small
\begin{tabular}{lccc}
\toprule
\multicolumn{4}{l}{\textbf{Panel A. Where does the label signal live?}} \\
\midrule
\textbf{Probe input} & & \textbf{AUROC} & \textbf{AUPRC} \\
\midrule
$U_T$ (notes) & & 0.8118 & 0.4175 \\
$V$ (expert summary) & & 0.8293 & 0.4788 \\
$\hat{V}$ (note-explainable part of $V$) & & 0.8157 & 0.4228 \\
$V_\perp$ (note-orthogonal part of $V$) & & 0.6906 & 0.2565 \\
$[U_T;V]$ & & 0.8339 & 0.4875 \\
\midrule
\multicolumn{4}{l}{\textbf{Panel B. Retrained ablation of the expert branch}} \\
\midrule
\textbf{Arm} & \textbf{AUPRC} & \textbf{$\Delta$AUROC (95\% CI)} & \textbf{$\Delta$AUPRC (95\% CI)} \\
\midrule
Physiology $+$ notes (reference) & 0.4054 & -- & -- \\
$+\ V$ (standardized) & 0.4977 & $+$0.043 [$+$0.029, $+$0.058] & $+$0.093 [$+$0.065, $+$0.121] \\
$+\ V_\perp$ (residual) & 0.4308 & $+$0.010 [$-$0.001, $+$0.020] & $+$0.026 [$+$0.005, $+$0.047] \\
$+\ V$ shuffled across patients & 0.3831 & $-$0.014 [$-$0.021, $-$0.008] & $-$0.022 [$-$0.036, $-$0.008] \\
$V_\perp$ only & 0.2445 & $-$0.115 [$-$0.145, $-$0.085] & $-$0.160 [$-$0.206, $-$0.115] \\
\bottomrule
\end{tabular}
\end{table}

\textbf{Ablation.} Substituting $V_\perp$ for the expert branch still improved AUPRC over the physiology-plus-notes reference by 0.0258 (95\% CI 0.005--0.047), so the benefit survives removing the ridge-predicted note component. The AUROC delta for the same arm included zero, concentrating the effect in the precision-recall regime that matters for an imbalanced outcome. The patient-shuffled control \emph{degraded} the reference by 0.0219 AUPRC, so the contribution is patient-specific, not a scale artifact.

The magnitudes are the substance of the finding. The standardized summary embedding is worth $+0.0928$ AUPRC over the reference; the residual alone is worth $+0.0258$, 28\% of that (22\% on AUROC). Read with the probe results, the summary's contribution is predominantly reorganization of information the notes already contain into a form the encoder and linear head can use. Because the two arms are separately retrained models whose gains need not sum, this is an indication of balance, not a decomposition.

\vspace{8pt}

\noindent\textbf{Discussion}

\vspace{4pt}

The framework's core proposition holds: encoding a structured summary yields a more useful representation than encoding the notes directly, and adds value over physiology and notes together. The mechanism, however, appears more mundane than knowledge injection. A 48-hour ICU record spans a median of 8 notes and roughly 9,700 characters, of which a 512-token encoder sees a fraction per note, mean-pooled into one vector; the LLM compresses that record into 264 words under fixed clinical headings, a far better match for what the encoder can consume. Much of the gain is a lossy channel being fed a better-prepared input, which relocates the target for improvement to the narrative-to-encoder interface.

\vspace{4pt}

\noindent\textit{Limitations}

\vspace{4pt}

Recoverability was tested with a \emph{linear} map, so 40.8\% is a lower bound on redundancy. Moreover $\alpha$ was selected for predictive $R^2$ rather than orthogonality, so the shrunk ridge leaves note-predictable variance inside $V_\perp$: the residual is not exactly orthogonal to $U_T$, and part of its $+0.0258$ may be note content the ridge missed. Symmetrically, $V_\perp$ discards any note signal the LLM cleaned up or re-weighted, making $+0.0258$ a lower bound on $V$'s value. The analysis is conditioned on Bio\_ClinicalBERT.

All arms were trained once at a fixed seed, so the intervals reflect test-set sampling variance only, not training-run variance; the residual arm's gain is small enough that seed variation could account for a meaningful share of it, and no comparison is corrected for multiplicity. It is provisional pending multi-seed replication. A logistic probe on $U_T$ also reached AUROC 0.8118, well above the 0.7473 of the trained notes-only network, so the note branch under-performs its own inputs and the measured expert contribution is relative to a weak reference. Input budgets are asymmetric (32 notes at 512 tokens versus a 4,096-token whole-record view), so part of $V$'s advantage may reflect coverage.

We report discrimination only; calibration was not assessed and is required before any deployment claim. We ran a single locally hosted LLM at temperature 0 and did not audit compliance with the no-prognostication constraint; because ``clinical gestalt'' may elicit prognostic framing, residual outcome-directed language cannot be excluded. Stays excluded for lack of a usable note had higher mortality (19.1\% vs 12.8\%), so the cohort under-represents patients with sparse early documentation. Absolute values are not directly comparable to published MIMIC-III results given these leakage controls and minimal fusion. Subgroup metrics improved in every group but carry no intervals and support no equity claim: the Black/African American subgroup (n=277, 27 deaths) gained 0.004 AUROC versus 0.044 for White patients, and the Hispanic subgroup's physiology-only AUROC was 0.532 (chance) on 9 deaths (Supplementary Table S1). Finally, this is one institution and era, and concatenative fusion precludes per-modality attribution.

\vspace{8pt}

\noindent\textbf{Conclusion}

\vspace{4pt}

LLM-generated expert summaries improve IHM prediction on MIMIC-III under a protocol with explicit leakage controls, and the gain is patient-specific and survives removal of the component linearly predicted from the note embedding. Most of it, however, comes from reorganizing information the notes already contain rather than from knowledge the LLM supplies. Framing the LLM as a representation-preparation step rather than a knowledge source is more defensible and more actionable, identifying the narrative-to-encoder interface as the component worth optimizing.

\vspace{8pt}

\noindent\textbf{Author Contributions}

\vspace{4pt}

HB designed the study, implemented the pipeline and analyses, and drafted the manuscript. JL contributed to experimental design and analysis. JS supervised the work. All authors reviewed and approved the final manuscript.

\vspace{8pt}

\noindent\textbf{Ethics Approval}

\vspace{4pt}

MIMIC-III is a de-identified, publicly available database, and its use is governed by the PhysioNet data use agreement. All expert summaries were generated by a model self-hosted on local NVIDIA H100 GPUs within the study's own compute environment: no patient-derived text was transmitted to any external or third-party inference service, so the considerations that apply to online LLM services do not arise.\textsuperscript{\ref{gptresponsibleuse}} De-identification placeholders were preserved in all prompts and no generated text left the environment. Deployments that substitute a hosted endpoint must independently confirm its permitted status. The framework is model-agnostic: summaries are cached by model ID and prompt hash, so changing the model changes one configuration field.

\vspace{8pt}

\noindent\textbf{Funding}

\vspace{4pt}

This research received no specific grant from any funding agency in the public, commercial, or not-for-profit sectors.

\vspace{8pt}

\noindent\textbf{Conflicts of Interest}

\vspace{4pt}

The authors declare no competing interests.

\vspace{8pt}

\noindent\textbf{Data Availability}

\vspace{4pt}
MIMIC-III is available to credentialed users through PhysioNet (\url{https://physionet.org}).\textsuperscript{\ref{johnson2016mimic}} The reference implementation that produced every figure and table is available at \url{https://github.com/HarBatt/MultiRep-IHM-Prediction}; each table corresponds to a cached artifact written by the analysis notebook.

\vspace{8pt}

\noindent\textbf{Acknowledgments}

\vspace{4pt}

We thank the MIT Laboratory for Computational Physiology for access to MIMIC-III.

\vspace{8pt}
\begin{center}
\textbf{References}
\end{center}

\vspace{4pt}
\begin{enumerate}
\renewcommand{\labelenumi}{\arabic{enumi}.}
\setlength{\itemsep}{0pt}
\setlength{\parskip}{0pt}
\small\setlength{\baselineskip}{9.6pt}
\item \label{harutyunyan2019multitask} Harutyunyan H, Khachatrian H, Kale DC, Ver Steeg G, Galstyan A. Multitask learning and benchmarking with clinical time series data. \textit{Sci Data.} 2019;6:96.
\item \label{ghassemi2015multivariate} Ghassemi M, Pimentel MAF, Naumann T, et al. A multivariate timeseries modeling approach to severity of illness assessment and forecasting in ICU with sparse, heterogeneous clinical data. \textit{Proc AAAI Conf Artif Intell.} 2015:446-453.
\item \label{suresh2018learning} Suresh H, Gong JJ, Guttag JV. Learning tasks for multitask learning: heterogenous patient populations in the ICU. \textit{Proc ACM SIGKDD Int Conf Knowl Discov Data Min.} 2018:802-810.
\item \label{song2018attend} Song H, Rajan D, Thiagarajan JJ, Spanias A. Attend and diagnose: clinical time series analysis using attention models. \textit{Proc AAAI Conf Artif Intell.} 2018:4091-4098.
\item \label{alsentzer2019publicly} Alsentzer E, Murphy JR, Boag W, et al. Publicly available clinical BERT embeddings. \textit{Proc 2nd Clin Nat Lang Process Workshop.} 2019:72-78.
\item \label{lee2020biobert} Lee J, Yoon W, Kim S, et al. BioBERT: a pre-trained biomedical language representation model for biomedical text mining. \textit{Bioinformatics.} 2020;36(4):1234-1240.
\item \label{bommasani2022foundation} Bommasani R, Hudson DA, Adeli E, et al. On the opportunities and risks of foundation models. \textit{arXiv.} 2021:2108.07258.
\item \label{tian2023justask} Tian K, Mitchell E, Zhou A, et al. Just ask for calibration: strategies for eliciting calibrated confidence scores from language models fine-tuned with human feedback. \textit{Proc Conf Empir Methods Nat Lang Process.} 2023:5433-5442.
\item \label{johnson2016mimic} Johnson AEW, Pollard TJ, Shen L, et al. MIMIC-III, a freely accessible critical care database. \textit{Sci Data.} 2016;3:160035.
\item \label{hochreiter1997long} Hochreiter S, Schmidhuber J. Long short-term memory. \textit{Neural Comput.} 1997;9(8):1735-1780.
\item \label{gptresponsibleuse} PhysioNet. Responsible use of MIMIC data with online services like GPT. \url{https://physionet.org/news/post/gpt-responsible-use}. Accessed 2026.
\end{enumerate}

\end{document}